\documentclass{article} 
\usepackage{iclr2023_conference,times}

\usepackage{hyperref}
\usepackage{url}

\usepackage{caption}
\usepackage{float}
\usepackage{subcaption}
\usepackage{booktabs}
\usepackage[pdftex]{graphicx}
\usepackage{multirow} 

\title{StrucTexTv3: An Efficient Vision-Language Model for Text-rich Image Perception, Comprehension, and Beyond}
\iclrfinalcopy
\hypersetup{hidelinks}

\author{Pengyuan Lyu\textsuperscript{$^\dag$}, Yulin Li\textsuperscript{$^\dag$}, Hao Zhou, Weihong Ma, Xingyu Wan, Qunyi Xie, \\ \textbf{Liang Wu, Chengquan Zhang\textsuperscript{$^*$}\thanks{$^\dag$Equal contribution, $^*$Corresponding Author.}, Kun Yao, Errui Ding, Jingdong Wang} \\
Department of Computer Vision Technology (VIS), Baidu Inc.\\
}
\makeatletter
\def\thanks#1{\protected@xdef\@thanks{\@thanks
        \protect\footnotetext{#1}}}
\makeatother
\begin{document}

\maketitle

\begin{abstract}
Text-rich images have significant and extensive value, deeply integrated into various aspects of human life. Notably, both visual cues and linguistic symbols in text-rich images play crucial roles in information transmission but are accompanied by diverse challenges. Therefore, the efficient and effective understanding of text-rich images is a crucial litmus test for the capability of Vision-Language Models. We have crafted an efficient vision-language model, \textbf{StrucTexTv3}, tailored to tackle various intelligent tasks for text-rich images. The significant design of StrucTexTv3 is presented in the following aspects:
\emph{Firstly}, we adopt a combination of an effective multi-scale reduced visual transformer and a multi-granularity token sampler (\textbf{MG-Sampler}) as a visual token generator, successfully solving the challenges of high-resolution input, like 1600x1600 pixels, and complex representation learning for text-rich images.
\emph{Secondly}, we enhance the perception and comprehension abilities of StrucTexTv3 through instruction learning, seamlessly integrating various text-oriented tasks into a unified framework.
\emph{Thirdly}, we have curated a comprehensive collection of high-quality text-rich images, abbreviated as \textbf{TIM-30M}, encompassing diverse scenarios like incidental scenes, office documents, web pages, and screenshots, thereby improving the robustness of our model. Our method achieved SOTA results in text-rich image perception tasks, and significantly improved performance in comprehension tasks.
Among multimodal models with LLM decoder of approximately \textbf{1.8B} parameters, it stands out as a leader, which also makes the deployment of edge devices feasible. In summary, the StrucTexTv3 model, featuring efficient structural design, outstanding performance, and broad adaptability, offers robust support for diverse intelligent application tasks involving text-rich images, thus exhibiting immense potential for widespread application.
\emph{Demo or API is coming soon.}

\end{abstract}

\section{Introduction}
Text-rich images serve as an important medium for conveying information. The extraction and comprehension of the embedded textual and visual data within them carry significant practical implications~\citep{huang2022layoutlmv3,yu2023structextv2,liu2024textmonkey,dong2024internlm4k}. This intricate process can be delineated into two primary capability dimensions: perception and comprehension. At the perceptual level, it primarily involves tasks related to visual perception, encompassing text spotting~\citep{wu2022decoupling,liu2023spts}, document parsing~\citep{blecher2023nougat,lv2023kosmos}, table recognition~\citep{lyu2023gridformer}, chart parsing~\citep{liu2022deplot}, among others. Conversely, the comprehension level pertains to abilities dependent on deeper semantic comprehension, including key information extraction, document-oriented VQA tasks (docvqa~\citep{mathew2021docvqa}, infovqa~\citep{mathew2022infographicvqa}, chartqa~\citep{masry-etal-2022-chartqa}, table image understanding~\citep{svetlichnaya2020deepform,zhao2023qtsumm}), text image translation, and so on.

\begin{figure}[H]
 \centering
 \includegraphics[width=1.0\textwidth]{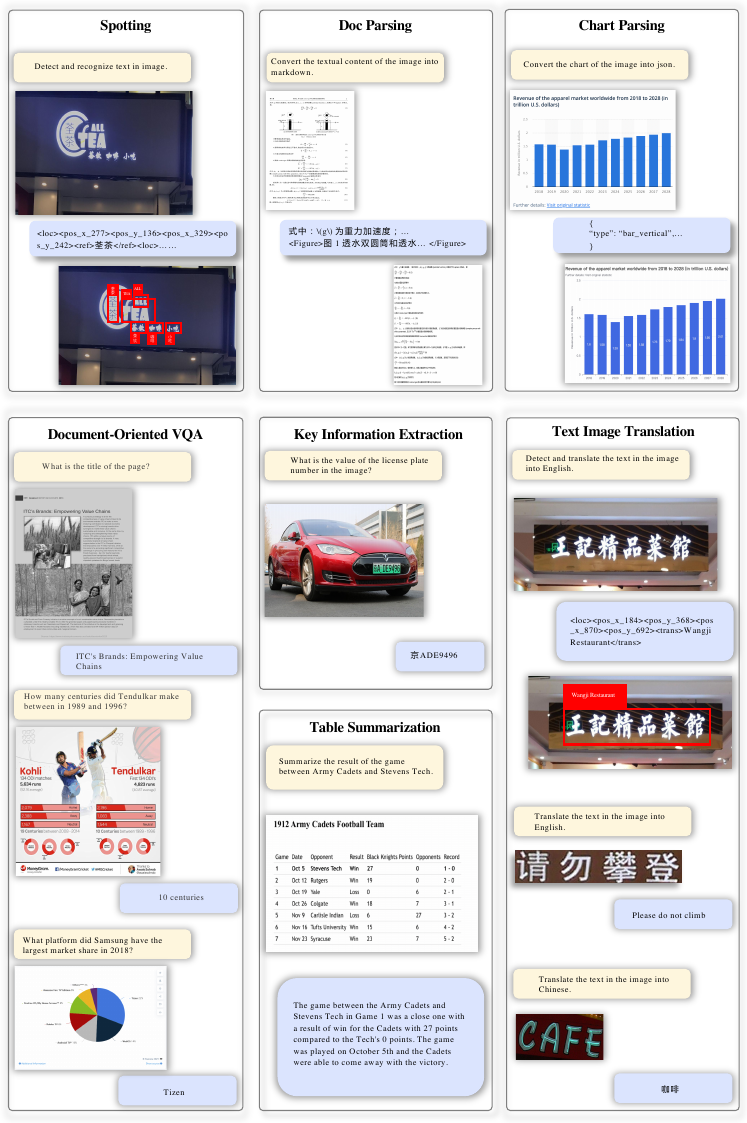}
\caption{General text-rich image perception and comprehension capabilities of StrucTexTv3. The first row displays perceptual-level capabilities such as text spotting, document parsing, and chart parsing. The second row presents cognitive-level abilities, including document-oriented VQA, key information extraction, table summarization, and text image translation.}
\label{fig:method_exps}
\vspace{-0.5em}
\end{figure}


In the early stages, the processing of perception and comprehension tasks often involved the utilization of multiple expert models and carried out in a multi-stage manner. However, in recent years, with the rapid advancement of multimodal large language models~\citep{lv2023kosmos,wei2023vary,liu2024textmonkey,dong2024internlm4k}, the development of an end-to-end universal model for handling perception and comprehension tasks has become a focal point of research. These approaches typically leverage pre-trained visual backbones to extract robust visual representations and combine them with the powerful comprehension capabilities of large-scale language models for semantic reasoning. Remarkably, these methods have demonstrated significant efficacy across multiple tasks.

Text-rich images, with their inherent diversity, complexity, and distinct perception-comprehension requirements, present a range of obstacles for multimodal large language models. A significant hurdle is the prevalence of small and dense text within these images, demanding high-resolution inputs for precise text extraction~\citep{li2023monkey}. Methods like LLaVA~\citep{liu2024visual} and Qwen-VL~\citep{bai2023qwen}, which commonly work with 224x224 or 336x336 image sizes, struggle to adequately capture this fine detail. To address this issue, some methods~\citep{DBLP:palix, DBLP:pali3} have attempted to use higher-resolution images directly. However, this approach is constrained by memory and computational resources, as the resolution that ViT-based visual encoders can handle is severely limited. Additionally, some methods~\citep{li2023monkey,ye2023ureader} explore the use of sliding window techniques to partition input images into smaller patches. While this approach partially mitigates the scale issue, it may introduce semantic incoherence, thereby affecting the perception and comprehension capabilities of the model.

In this paper, we introduce \textbf{StrucTexTv3}, an efficient vision-language model designed for the perception and comprehension of text-rich images. 
Diverging from previous approaches that employ ViT as the image encoder, StrucTexTv3 adopts a hierarchical vision transformer (such as Swin Transformer). 
This design ensures high-resolution inputs, such as 1600x1600, while mitigating semantic inconsistencies and minimizing memory usage. 
Additionally, to acquire robust visual representations capable of addressing diverse perception or comprehension tasks, we introduce a multi-granularity token sampler (abbreviated as \textbf{MG-Sampler}). 
This sampler extracts rich visual representations from multi-scale feature spaces, ensuring high-quality and comprehensive visual representations. Building upon this efficient model, we have integrated a variety of tasks tailored for text-rich images using instruction learning. These tasks include perception tasks such as text spotting and document parsing, as well as comprehension tasks like documented-oriented vqa, key information extraction, table image understanding, and text image translation. In addition, to ensure the robustness of our model, we have established a data pipeline for generating high-quality perception and comprehension data, collecting nearly 30 million of \textbf{T}ext-rich \textbf{I}mage based \textbf{M}ultimodal data (abbreviated as \textbf{TIM-30M}).


We validated our model across multiple benchmarks, and the experimental results reveal remarkable performance despite its modest \textbf{1.8B LLM}, outperforming models with \textbf{7B LLM} across various tasks and datasets. 

Our contributions can be summarized as follows:
\begin{itemize}

\item We introduce an efficient vision-language model (StrucTexTv3) tailored for the perception and comprehension of text-rich images, adeptly tackling challenges associated with high-resolution input and intricate representation learning.

\item By employing multiple data collection methods, we have curated a comprehensive collection of high-quality text-rich images named TIM-30M, which includes rich perceptual and comprehension instruction learning data.

\item Employing instruction learning, StrucTexT v3 demonstrates extensive perception and comprehension capabilities, culminating in state-of-the-art performance across numerous tasks and benchmark datasets.

\end{itemize}

\section{Model Architecture}
\begin{figure*}[ht]
 \centering
 \includegraphics[width=0.8\textwidth]{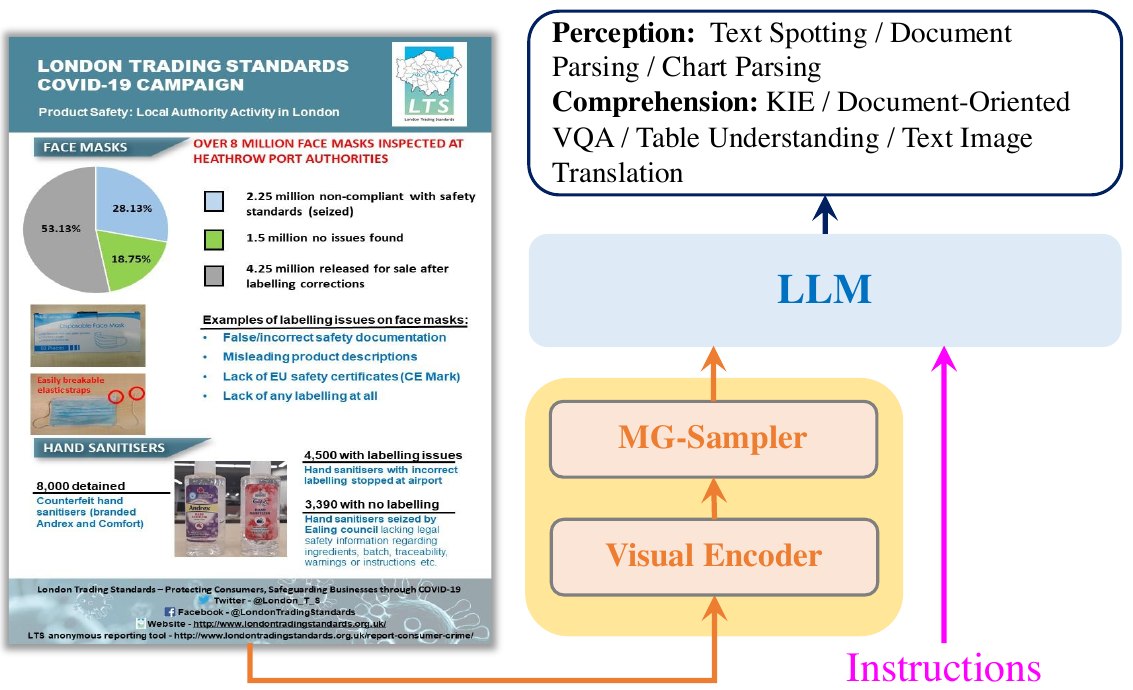}
\caption{The overview of StrucTexTv3. It comprises three core parts: Visual Encoder, MG-Sampler, and LLM.}
\label{fig:method_arch}
\vspace{-0.5em}
\end{figure*}
StrucTexT v3 comprises three components: (1) an efficient \textbf{V}isual \textbf{E}ncoder for effective extraction of visual features, (2) a \textbf{M}ulti-\textbf{G}ranularity token \textbf{S}ampler for deriving rich fine-grained visual representations from multi-scale visual features, (3) a \textbf{L}arge \textbf{L}nguage \textbf{M}odel for understanding and reasoning about visual cues and linguistic symbols in text-rich images.

\textbf{Visual Encoder.}  We employed a Swin-Large model~\citep{liu2021swin} that has been pretrained on ImageNet~\citep{deng2009imagenet} as our visual encoder to extract hierarchical visual representations. The intrinsic shifted windowing scheme of the Swin Transformer ensures efficiency even when handling higher resolution inputs. Leveraging this characteristic, we configure the resolution of the input images to 1600x1600 pixels, thereby facilitating the accurate detection and recognition of even small-scale text.

\textbf{MG-Sampler.} We employed a resampler with a structure akin to Q-Former~\citep{li2023blip} to extract refined visual representations from lengthy, redundant visual representations. Additionally, given that the Swin Transformer has a hierarchical architecture, we can extract features at various levels of granularity from the outputs of different stages using the resampler. Specifically, we extracted feature representations with a sequence length of 256 from the outputs of Stage 3 and Stage 4 of the Swin Transformer. These two feature sequences were then concatenated into a single feature sequence of length 512, which served as the input for the language model.

\textbf{LLM.} We adopted the standard transformer decoder structure~\citep{vaswani2017attention} as our language model and have been effectively trained on a large amount of Chinese and English web-page text, codes, book contents, and other data. 
To ensure edge device compatibility, our optimized LLM boasts a parameter count of 1.8B, featuring 24 layers, 16 attention heads, and 2048 hiddens. 
To simplify the encoding length of two-dimensional coordinate sequences, we have added two sets of tokens with an accuracy value of one thousandth for relative x/y coordinates, and the overall vocabulary size is approximately 160,000.

\section{Data Construction and Prompt Design}
Rich and diverse data are crucial for ensuring that large multimodal models acquire perception and comprehension capabilities. Based on this premise, we have constructed a large-scale text-rich image based multimodal training data, namely \textbf{TIM-30M}, which mainly includes perception tasks and comprehension tasks. Excluding some business-owned data, we will provide a detailed introduction to the composition of these two data types of capability (as listed in Table ~\ref{table_training}), especially the public training data included and the corresponding prompt design formats.

\subsection{Perception Data}
We have curated a comprehensive, large-scale training dataset that focused on enhancing perception abilities in areas like text spotting, document parsing (including normal text, formula, and table), and chart parsing. The accumulated training data boasts a total of approximately 25 million text-rich images. 

\textbf{Spotting.}
We construct a large-scale spotting training dataset involving publicly available benchmarks, synthetic data, and pseudo-labeled in-house data to endow the model with robust spotting capabilities. The composition of the benchmarks is presented in Table \ref{table_training}, which contains  ICDAR2013~\citep{DBLP:icdar2013}, ICDAR2015~\citep{karatzas2015icdar},
MLT2017~\citep{DBLP:mlt}, TextOCR~\citep{DBLP:textocr}, CTW1500~\citep{DBLP:ctw}, TotalText~\citep{DBLP:total_text}, HierText~\citep{DBLP:conf/cvpr/LongQPBFR22}, CORD~\citep{park2019cord}, FUNSD~\citep{jaume2019funsd}, SROIE~\citep{huang2019icdar2019},
RECTs~\citep{DBLP:rects}, LSVT~\citep{DBLP:lsvt} and HWDB~\citep{DBLP:hwdb}. For all spotting data, we used ``Detect and recognize text in image" as the prompt. We represent each text instance as a sequence of characters formatted as \texttt{\textless{}pos\_x1\textgreater{}\textless{}pos\_y1\textgreater{}\textless{}pos\_x2\textgreater{}\textless{}pos\_y2\textgreater{}\textless{}ref\textgreater{}transcription\textless{}/ref\textgreater{}}. Here, \texttt{\textless{}pos\_x1\textgreater{}\textless{}pos\_y1\textgreater{}\textless{}pos\_x2\textgreater{}\textless{}pos\_y2\textgreater{}} are specialized tokens denoting the coordinates of the top-left and bottom-right vertices of the text. The string enclosed between \texttt{\textless{}ref\textgreater{}} and \texttt{\textless{}/ref\textgreater{}} represents the content of the text. In our experiments, we found that there was no significant difference in results between representing each coordinate as a sequence of characters using multiple tokens and directly using a special token for each coordinate. However, the special token representation significantly reduces the length of the generated sequence, particularly for samples with a large number of text instances, thereby greatly reducing the inference overhead. Subsequently, we sorted all text instances based on their positions from top to bottom and from left to right and generated labels accordingly. We found that organizing the text in this manner reduces label ambiguity and consequently enhances the performance of our model.

\textbf{Document Parsing.} We also build a large-scale training dataset for document parsing. Our data sources are primarily composed of three parts:  publicly available benchmarks, arXiv data, and a weakly labeled in-house dataset. We collect FinTabNet~\citep{DBLP:fintabNet} and PubTabNet~\citep{DBLP:pubtabnet} for table structure recognition to enable our model to parse table structures. Each table image is represented as an HTML sequence. We follow the data production process outlined by \cite{blecher2023nougat}, downloading LaTeX source code from arXiv, rendering it into PDF format, and converting it into corresponding Markdown sequences. We have also amassed a substantial collection of books and academic paper images, and have utilized our commercial API \footnote{\url{https://ai.baidu.com/tech/ocr/doc_analysis_office}} to acquire pseudo-labels to enhance the generalization capabilities of our model. We use the prompt ``Convert the textual content of the image into markdown." to instruct the model to parse the input into Markdown format.

\textbf{Chart Parsing.}
Due to the limited amount of data available in public benchmarks for chart parsing, in addition to collecting data from public benchmarks including ChartQA~\citep{masry-etal-2022-chartqa}, PlotQA~\citep{DBLP:plotqa}, Chart-to-Text~\citep{DBLP:chart_to_text}, UniChart~\citep{DBLP:unichart} and  MMC~\citep{liu2023mmc}, 
we also developed a synthetic data generation pipeline to generate Chart images with several visulization packages (\emph{e.g.}, ECharts\footnote{ \url{https://echarts.apache.org/en/index.html}}, 
Matplotlib\footnote{\url{https://matplotlib.org}}, 
plotly\footnote{\url{https://github.com/plotly/plotly.py}}, etc). 
The pipeline supports the creation of common chart types such as histograms, bar charts, pie charts, mind map and so on. 
To enrich the instruction data, Ernie-Bot-4.0~\footnote{\url{https://qianfan.cloud.baidu.com}} is adopted for generating statistical data, QA pairs, and summaries.
To guide the model in transforming chart images into sequences in a target format, we use the format ``Convert the chart of the image into \{target format\}". Here, \{target format\} refers to the sequence format, which can be CSV, Markdown, or JSON. This approach enhances the model’s capability to process various chart types and improves its accuracy in converting visual data into structured data.

\subsection{comprehension Data}
The comprehension capabilities of our model are inherently linked to various tasks, encompassing key information extraction, document-oriented VQA, table image understanding, and text image translation. The total number of associated text-rich images is 5 million.

\textbf{Key Information Extraction.}
We collect open-source datasets including CORD~\citep{park2019cord}, EPHOIE~\citep{WangLJT0ZWWC21}, SROIE~\citep{huang2019icdar2019}, XFUND~\citep{XuL0WLFZW22}, SIBR~\citep{yang2023modeling}, WildReceipt~\citep{sun2021spatial}, Form-NLU~\citep{ding2023form}, Kleister-NDA~\citep{StanislawekGWLK21} and Chinese in-house data to enrich the model's power for both Chinese and English KIE tasks. A brief prompt template: ``What is the value of the \{key\}?" is applied for semantic entity extraction. Note that for several manual labels or abbreviated phrases, such as \textit{void\_menu.nm} in CORD, \textit{address\_\_street\_line} in Kleister-NDA, we imitate the practices of InstructDoc~\citep{TanakaINSS24} that construct prompts holding all category options and constrain the model to select one of them.

\textbf{Document-Oriented VQA.}
In this work, we utilize a diverse set of OCR-related VQA datasets to develop better visual document comprehension in scene text scenarios. The extensive datasetes consists of 13 pieces: DocVQA~\citep{mathew2021docvqa}, InfoVQA~\citep{mathew2022infographicvqa}, DUDE~\citep{LandeghemPTJBBC23}, TextOCR~\citep{DBLP:textocr}, SlideVQA~\citep{TanakaNNHSS23}, WebSRC~\citep{ChenZCJZLX021}, TextbookQA~\citep{KembhaviSSCFH17}, VisualMRC~\citep{TanakaNY21}, TabFact~\citep{ChenWCZWLZW20}, BenthamQA~\citep{MathewGKJ21}, ST-VQA~\citep{BitenTMBRJVK19}, OCR-VQA~\citep{0001SSC19}, PDF-VQA~\citep{DingLCH23}. Additionally, we use the self-instruction method~\citep{li2023blip} to expand the diversity of data and supplement the LLaVA-150k~\citep{liu2024visual} instruction data to enrich the general conversational capabilities of our model.

\textbf{Table Image Understanding.}
We enhance our model’s capability to comprehend tables by utilizing two tasks: Table VQA and Table Summarization. The training data of table image understanding involves publicly available benchmarks, and pesudo-labeled in-house data.
The composition of the benchmarks is presented in Table \ref{table_training}, which contains QTSUMM~\citep{zhao2023qtsumm}, ToTTo~\citep{parikh2020totto}, LogicNLG~\citep{chen2020logical}, Deepform~\citep{svetlichnaya2020deepform}, 
 and WTQ~\citep{pasupat2015compositional}. Each sample’s question is directly used as a prompt. To enhance the diversity of our training data and to mitigate the overfitting issue due to the insufficient volume of training samples in the public table image understanding benchmarks, we also employed Ernie-Bot-4.0 to generate QA pairs for data augmentation.

\textbf{Text Image Translation.}
We also explore the translation capabilities of our model by constructing an image-to-text translation dataset. Our data sources include publicly available benchmark OCRMT30K~\cite{DBLP:ocrmt30k} and synthetic data. The data encompasses both entire images and text line images. For text-line images, we use ``Translate the text in the image into English/Chinese." as the prompt. For entire images, we employ prompts such as ``Detect and translate the text in the image into English/Chinese." and ``Detect, recognize, and translate the text in the image into English/Chinese.". The text instances are still organized from top-left to right-bottom.
\begin{table}[ht]
\caption{Summary of \textbf{TIM-30M} data.}
\label{table_training}
\vspace{-0.5em}
\begin{center}
\begin{tabular}{lll}
\toprule[1pt]
\textbf{Capability} & \textbf{Task} & \textbf{Dataset}  \\
\midrule
\multirow{10}{*}{Perception} & 
\multirow{3}{*}{Spotting} & \parbox{6.5cm}{Public data: MLT2017, ICDAR2013, ICDAR2015, TextOCR, CTW1500, Total-Text, HierText, CORD, FUNSD, SROIE, RECTs, LSVT, HWDB. }   \\
~& ~ & Synthetic data.  \\
~& ~ & In-house data. \\
\cmidrule{2-3}
~ & 
\multirow{3}{*}{Document parsing} & \parbox{6.56cm}{Public data: FinTabNet, PubTabNet}  \\
~ & ~ & arXiv data.  \\ 
~ & ~ & In-house data  \\
\cmidrule{2-3}
~ &
\multirow{2}{*}{Chart parsing} & \parbox{6.5cm}{Public data: ChartQA, PlotQA, UniChart, Chart-to-Text.}  \\
~ & ~ & Synthetic data. \\

\midrule
\multirow{12}{*}{Comprehension} &
\multirow{2}{*}{Key information extraction} &\parbox{6.5cm} {Public data: CORD, EPHOIE, SROIE, SIBR, XFUND, WildReceipt, Form-NLU, Kleister-NDA.}  \\
~ & ~ & In-house data.\\
\cmidrule{2-3}
~ & 
\multirow{2}{*}{Document-Oriented VQA} & \parbox{6.5cm}{Public data: DocVQA, WebSRC, DUDE, InfoVQA, TextOCR, TabFact, OCR-VQA, VisualMRC, ST-VQA, SlideVQA, BenthamQA, TextbookQA, PDF-VQA, LLaVA-150k.} \\
~ & ~ & In-house data. \\
\cmidrule{2-3}
~ &
\multirow{2}{*}{Table image understanding} & \parbox{6.5cm}{Public data: QTSUMM, Totto, LogicNLG, DeepForm, WTQ.}  \\
~ & ~ & In-house data. \\

\cmidrule{2-3}
~ &
\multirow{2}{*}{Text image translation} & Public data: OCRMT30K \\
~& ~ & Sythetic data.   \\

\bottomrule[1pt]
\end{tabular}
\end{center}
\vspace{-0.5em}
\end{table}

\section{Training Pipelines}
Our training process includes three stages: pre-training, multi-task pre-training, and supervised fine-tuning. Following~\citep{bai2023qwen,DBLP:deepseek-vl}, we mixed image-text pair data with pure-text data during training to maintain the capabilities of the language model.

\textbf{Pre-training.} 
During the pre-training phase, our objective is to endow our model with the capability to perceive text. To reduce training difficulty, we focus solely on training the spotting task. All parameters are trained unfrozen with a sequence length of 4096. To expedite the training process, the input image size was initially 960 and later increased to 1600.

\textbf{Multi-task Pre-training.}
In this stage, we utilize all training data from multiple perceptual and cognitive tasks to support our model with both perceptual and cognitive capabilities. The size of the input images remains at 1600x1600, and the maximum sequence length remains at 4096. Additionally, all parameters are trainable.

\textbf{Supervised Fine-tuning.}
Due to the large amount of synthetic or pseudo-labeled data in our training set, we further fine-tuned our model using the relatively high-quality benchmark data listed in Table~\ref{table_training} during the third phase. We retained the same image size, sequence length, and trainable parameters as in the second phase, additionally training for 2000 steps on higher-quality benchmarks.

\section{Evaluation}
In this section, we assess the perception and comprehension abilities of StrucTexTv3. The perception assessment focuses on text spotting, document parsing, and chart parsing. Meanwhile, the comprehension evaluation includes benchmarks for key information extraction, document-oriented VQA, table image understanding, and text image translation.

\subsection{Perception Ability}

\subsubsection{Text Spotting}


The text spotting task is evaluated on two benchmark datasets, the document image dataset SROIE~\citep{huang2019icdar2019} and the scene image dataset ICDAR15~\citep{karatzas2015icdar}, respectively. To facilitate comparison with previous methods, we employed the official evaluation protocol for SROIE, assessing word-level precision, recall, and F1 score. For ICDAR2015, we utilized the evaluation scripts provided by SPTS~\citep{liu2023spts}, using both point-based and transcription-based metrics to evaluate the results. Given methods without publicly available results or a unified evaluation protocol on the corresponding benchmark datasets, we conducted fair evaluations based on the officially released implementations.

As shown in Tab.~\ref{tab_spotting}, our method achieves the best performance on both the SROIE and ICDAR2015 datasets. On the SROIE dataset, we obtain 94.1\% f1 score, as for the ICDAR2015, the metrics for Trans and pos are 62.4\% and 69.5\% correspondingly.



\begin{table}[h]
\caption{Experimental results on Text Spotting benchmarks. “*” denotes the results obtained through the open-source checkpoint model.}
\centering
\begin{tabular}{l c c c}
\toprule[1pt]
    \multirow{2}{*}{\textbf{Methods}} & \textbf{LLM Size} & \textbf{SROIE} & \textbf{ICDAR2015}\\ 
\cmidrule(rl){3-3}  \cmidrule(rl){4-4}
    ~ & ~ & \bf P / R / F1  &\bf Trans / Pos \\ 
\midrule

Vary~\citep{wei2023vary}$^{*}$  & 7B & 21.7 / 48.5 / 30.0 & -- / -- \\
TextMonkey~\citep{liu2024textmonkey}$^{*}$  & 7B &67.0 / 47.8 / 55.8 & 59.6 / 55.9 \\
\midrule
KOSMOS-2.5~\citep{lv2023kosmos}  & 0.85B &91.7 / 92.6 / 92.1 &  -- / -- \\
Ours & 1.8B & \textbf{94.2} / \textbf{94.1} / \textbf{94.1}  & \textbf{62.4}	/ \textbf{69.5} \\
\bottomrule[1pt]
\end{tabular}
 \vspace{0.2cm}
\label{tab_spotting}
\end{table}

\subsubsection{Document Parsing}
For document parsing, the model outputs the Markdown conversion results when provided with the document image, including the main text, table HTML information, and formula recognition results.
Following~\citep{wei2023vary}, we evaluate the performance of our model using three test sets on this task, comprising 200 document images for main text parsing, 100 document images with mathematical formulas, and 100 document images with span-cell tables. In each dataset, both Chinese and English images make up half of the data. Following ~\citep{lv2023kosmos}, we compare the 1-Normalized Edit Distance with other methods that have demonstrated decent performance in this task. When evaluating document images with formulas, some existing methods may treat main text, such as numbers, as formulas. 
Therefore, the main text is taken into account when evaluating formula performance.
On the other hand, when evaluating document images with tables, we exclusively evaluate the content within the tables to emphasize the effectiveness of table recognition.

As shown in Tab.~\ref{tab_doc}, when considering the output consisting solely of pure main text content, our method outperforms other competing methods in both Chinese and English documents. Furthermore, when considering the document image with formula and table recognition result, our method also achieves the best performance, significantly surpassing other comparative methods, showing impressive capabilities in perceiving and comprehending textual information.


\begin{table*}[h]
\caption{Experimental results on Doc Parsing benchmarks. “*” denotes the results obtained through the open-source checkpoint model. In formula set, formula sequence lengths accounts for 65\% of the total sequence in Chinese set, while 45\% in English set.}
\small
\centering
\setlength{\tabcolsep}{3.2pt}
{
\begin{tabular}{lccccccc}
\toprule[1pt]
\multirow{2}{*}{\textbf{Methods}} & \multirow{2}{*}{\textbf{LLM Size}} & \multicolumn{2}{c}{\textbf{Pure Document OCR}} & \multicolumn{2}{c}{\textbf{Formula}} & \multicolumn{2}{c}{\textbf{Table}} \\ 
\cmidrule(rl){3-4} \cmidrule(rl){5-6} \cmidrule(rl){7-8} & ~ & Chn(\%) & Eng(\%) & Chn(\%)  & Eng(\%) & Chn(\%)  & Eng(\%)  \\
\midrule
\multirow{1}{*}{Vary~\citep{wei2023vary}$^{*}$} 
&  7B & 37.6 &  50.8 &  31.3 &  61.9 &  57.8 & 2.4  \\ 
\midrule
\multirow{1}{*}{Nougat~\citep{blecher2023nougat}$^{*}$}  
&  0.27B &  3.6 &  78.8 &  5.4 &  86.2 &  0.0 & 71.7 \\ 
\multirow{1}{*}{Kosmos2.5~\citep{lv2023kosmos}$^{*}$} 
&  0.85B & 31.8 &  67.9 &  21.3 &  63.9 &  40.4 & 50.3  \\ 
\multirow{1}{*}{Vary-toy~\citep{wei2024small}$^{*}$} 
&  1.8B & 33.1 &  65.5 &  31.3 &  60.9 &  74.4 & 41.3  \\ 
\multirow{1}{*}{Ours} 
&  1.8B & \bf{95.1} &  \bf{96.9} &  \bf{92.7} &  \bf{95.7} &  \bf{82.5} & \bf{92.2}  \\ 
\bottomrule[1pt]
\end{tabular}
\label{tab_doc}
}
\end{table*}

\subsubsection{Chart Parsing}
We evaluate the model performance of Chart Parsing on the chart-to-table set of ChartQA~\citep{masry-etal-2022-chartqa}.
For chart-to-table translation, we use RMS-F1 from DePlot~\citep{liu2022deplot}. 
Few multimodel models master the ability of chart parsing. 
Hence, as shown in Tab~\ref{tab:chart_to_table}, we reported the performance of several methods specifically designed for chart-related tasks. 
Thanks to the optimization on diverse text-rich data, our method achieved the best results in the chart parsing task, surpassing several models with 13B LLM.

\begin{table*}[h]
\caption{Experimental results on Chart Parsing benchmarks. “*” denotes the results obtained through the open-source checkpoint model.}
\label{tab:chart_to_table}
\centering
\setlength{\tabcolsep}{9pt}
{
\begin{tabular}{l c c}
\toprule[1pt]
\multirow{2}{*}{\textbf{Methods}} & \multirow{2}{*}{\textbf{LLM Size}} & \multicolumn{1}{c}{\textbf{Chart-to-Table}}  \\ 
\cmidrule(rl){3-3} 
~ & ~ & RMS-F1   \\
\midrule 
ChartLlama~\citep{han2023chartllama}  & 13B   & 90.0 \\ 
ChartAst~\citep{meng2024chartassisstant}   & 13B & 91.6 \\
\midrule
OneChart~\citep{chen2024onechart} & 0.2B & ~88.8*  \\ 
Matcha~\citep{liu-etal-2023-matcha}   & 1.3B  & 89.6 \\ 
Ours       & 1.8B  &   \bf{91.8}     \\ 

\bottomrule[1pt]
\end{tabular}
}
\end{table*}

\subsection{Comprehension Ability}

\subsubsection{Key Information Extraction}
We evaluate the model performance of Key Information Extraction on three public benchmarks: FUNSD~\citep{jaume2019funsd}, POIE~\citep{KuangHLYJRB23}, and SVRD~\citep{YuZCHLCLCKCDFHLYYLCDLLYZKSWB23}.  For a fair comparison, we employ a zero-shot setup and remove the English documents of XFUND and the training examples of SVRD during training, to eliminate the risk of benchmark leakage. We follow the instruction templates released by InstructDoc and adopt the entity F1-score for the FUNSD labeling task. The accuracy metric~\citep{liu2024textmonkey} is utilized for measuring POIE and the FUNSD linking task. We mainly assess the models’ ability on task 3 of SVRD (E2E Zero-shot Structured Text Extraction) by the ICDAR official leaderboards. As shown in Table~\ref{kie}, our model shows comparable scores of FUNSD to that of 7B-size models and achieves state-of-the-art performance on POIE and SVRD. The results demonstrate that our model proves great comprehension ability with only 1.8B size of LLM, indicating its effectiveness in the KIE tasks.


\begin{table}[ht]
\caption{Experimental results on key information extraction. The superscript “*” refers to the results obtained through the open-source checkpoint model.}
\label{kie}
\vspace{-0.5em}
\begin{center}
\setlength{\tabcolsep}{4.5pt}
{
\begin{tabular}{l c c c c c}
\toprule[1pt]

\multirow{2}{*}{\textbf{Methods}} & \multirow{2}{*}{\textbf{LLM Size}} & \multicolumn{2}{c}{\textbf{FUNSD}} & \multicolumn{1}{c}{\textbf{POIE}} & \multicolumn{1}{c}{\textbf{SVRD}}  \\ 
\cmidrule(rl){3-4} \cmidrule(rl){5-5} \cmidrule(rl){6-6}
& & labeling (F1) & linking (Acc.) & Acc. & F1-score   \\
\midrule

Monkey~\citep{li2023monkey} & 7B & - & 24.1 & 19.9 & ~37.2$^{*}$ \\
DocPedia~\citep{Feng2023DocPediaUT} & 7B & - & 29.9 & 39.9 & - \\
TextMonkey~\citep{liu2024textmonkey} & 7B & - & \textbf{42.9} & 32.0 & ~51.4$^{*}$ \\
\midrule
InstructDoc~\citep{TanakaINSS24} & 3.4B & 38.2 & - & - & - \\
Ours & 1.8B & \textbf{55.7} & \textbf{42.9} & \textbf{55.9} & \textbf{55.6} \\
\bottomrule[1pt]
\end{tabular}
}
\end{center}
\vspace{-0.5em}
\end{table}

\subsubsection{Document-Oriented VQA}

In this study, our primary focus lies on Document-Oriented VQA, and we assess our model's performance across three benchmarks: DocVQA, InfoVQA, and ChartQA. DocVQA and InfoVQA employ ANLS (Average Normalized Levenshtein Similarity), an edit distance-based metric, for evaluation. For ChartQA, we adopt the methodology outlined in prior works~\citep{masry-etal-2022-chartqa,liu2022deplot}, allowing for relaxed correctness (an exact match with a tolerance for a 5\% numerical error). As illustrated in Table~\ref{tab:vqa}, our model achieves 72.8\% ANLS on DocVQA, 28.9\% ANLS on InfoVQA, and 70.8\% average accuracy on ChartQA, surpassing Vary-toy~\citep{wei2024small} which has a similar LLM size to ours, by a substantial margin. It underscores its adeptness in document understanding. However, our model demonstrates relatively poor performance on the challenging InfoVQA task, likely attributable to the limitations posed by its 1.8B LLM size on generalization ability.

\begin{table}[h]
\caption{Experimental results on Document-Oriented VQA benchmarks.}
\label{tab:vqa}
\vspace{-0.5em}
\begin{center}
\begin{tabular}{l c c c c}
\toprule[1pt]
\multirow{2}{*}{\textbf{Methods}} & \multirow{2}{*}{\textbf{LLM Size}} & \multicolumn{1}{c}{\textbf{DocVQA}} & \multicolumn{1}{c}{\textbf{InfoVQA}} & \multicolumn{1}{c}{\textbf{ChartQA}} \\ 
\cmidrule(rl){3-3} \cmidrule(rl){4-4} \cmidrule(rl){5-5} 
& & ANLS & ANLS & Average  \\
\midrule  
ChartLlama~\citep{han2023chartllama} & 13B & - & - & 69.6 \\ 
ChartAst~\citep{meng2024chartassisstant} & 13B & - & - & 75.1 \\
\midrule 
DocPedia~\citep{Feng2023DocPediaUT} & 7B & 47.1 & 15.2 & 46.9 \\
DocOwl~\citep{abs-2307-02499}  & 7B & 62.2 & 38.2 & 57.4 \\
Qwen-VL~\citep{bai2023qwen} & 7B & 65.1 & 35.4 & 65.7 \\ 
TextMonkey~\citep{liu2024textmonkey} & 7B & 73.0 & ~38.7$^{*}$ & 66.9 \\
HRVDA~\citep{liu2024hrvda} & 7B & 72.1 & 43.5 & 67.6 \\
DocOwl-1.5~\citep{hu2024mplugdocowl} & 7B &  81.6 & 50.4 & 70.5 \\
IXC2-4KHD~\citep{dong2024internlm4k} & 7B & \textbf{90.0} & \textbf{68.6} & \textbf{81.0}  \\
\midrule
InstructDoc~\citep{TanakaINSS24} & 3.4B & - & 50.9 & 29.4 \\
Vary-toy~\citep{wei2024small} & 1.8B  & 65.0 & ~6.3$^{*}$ & 59.1 \\
Matcha~\citep{liu-etal-2023-matcha} & 1.3B & 74.2 & 37.2 & 64.2 \\
Ours & 1.8B  & 72.8 & 28.9 & 70.8 \\ 
\bottomrule[1pt]
\end{tabular}
\end{center}
\vspace{-0.5em}
\end{table}

\subsubsection{Table Image Understanding}
Table image understanding aims to precise extraction and interpretation of tabular data to address user-defined queries effectively.
To further assess the performance of table image understanding, we conduct experiments on two tabular data-oriented datasets, specifically tailored for table VQA and table summarization tasks. For the benchmark evaluation of the table VQA task on Deepform~\citep{svetlichnaya2020deepform} dataset, we employed the normalized accuracy score metric. For the table summarization task, we utilized the METEOR~\citep{banerjee2005meteor} and AutoACU~\citep{liu2022revisiting} metrics on QTSUMM~\citep{zhao2023qtsumm}. We rendered tabular data into table images and filtered out the ones including multiple images (due to the single image input limitation). 
As shown in Tab.~\ref{table_understanding}, our method obtained the best performance on QTSUMM and the second best performance on Deepform. Compared to the previous SOTA method Qwen-VL~\citep{bai2023qwen}, the METEOR/AutoACU metrics on QTSUMM raised from 22.04\%/23.71\% to 31.81\%/36.04\% respectively. Our method also achieves comparable performance to SOTA method DocOwl-1.5~\citep{hu2024mplugdocowl} on the DeepForm dataset, while the LLM size is only 1.8B.


\begin{table}[ht]
\caption{Experimental results on table image understanding benchmarks. “*” denotes the results obtained through the open-source checkpoint model.}
\label{table_understanding}
\vspace{-0.5em}
\begin{center}
\begin{tabular}{l c c c c}
\toprule[1pt]
\multirow{2}{*}{\textbf{Methods}} & \multirow{2}{*}{\textbf{LLM Size}}  & \multicolumn{1}{c}{\textbf{Deepform}} & \multicolumn{2}{c}{\textbf{QTSUMM}} \\
\cmidrule(rl){3-3} \cmidrule(rl){4-5}
  & & Acc. & METEOR & AutoACU \\

\midrule

Qwen-VL~\citep{bai2023qwen} & 7B & 3.1	& 22.0$^{*}$ & 23.7$^{*}$ \\
Monkey~\citep{li2023monkey} & 7B & 40.5 & 2.6$^{*}$ & 10.4$^{*}$ \\
TextMonkey~\citep{liu2024textmonkey} & 7B & 61.6	& 2.4$^{*}$ & 10.8$^{*}$ \\
DocOwl-1.5~\citep{hu2024mplugdocowl} & 7B & \textbf{68.8} & 12.1$^{*}$ & 20.6$^{*}$ \\
\midrule
Ours & 1.8B & 65.8 &  \textbf{31.8} & \textbf{36.0} \\
\bottomrule[1pt]
\end{tabular}
\end{center}
\vspace{-0.5em}
\end{table}


\subsubsection{Text Image Translation}
We found that many existing multimodal large language models already possess the ability to accurately recognize text content in images. However, for the task of image translation, which requires both perception and understanding abilities, their performance is suboptimal. Therefore, we aimed to verify whether incorporating relevant training data during the training phase could enable the model to acquire image translation capabilities. We synthesized 1M training and 1000 testing data for image translation based on the WMT18 Chinese to English corpus and evaluated the model's performance. As shown in Table~\ref{image_translation}, compared to models that were not supervised-trained on the relevant task, our model demonstrated a significant improvement in image translation performance, demonstrating the potential of multimodal large language models in this task.

\begin{table}[ht]
\caption{Experimental results on image translation task. The superscript “*” refers to the results obtained through the open-source checkpoint model or API.}
\label{image_translation}
\vspace{-0.5em}
\begin{center}
\begin{tabular}{l c c }
\toprule[1pt]
\textbf{Methods} & \textbf{LLM Size}  & \textbf{sacre-BLEU} \\

\midrule

Qwen-VL~\citep{bai2023qwen}$^{*}$ & 7B & 0.72 \\
IXC2-4KHD~\citep{dong2024internlm4k}$^{*}$ & 7B & 1.86  \\
GPT 4o~\footnotemark$^{*}$ & - & \textbf{20.36}  \\
\midrule
Ours & 1.8B & 12.0 \\
\bottomrule[1pt]
\end{tabular}
\end{center}
\vspace{-0.5em}
\end{table}
\footnotetext{\url{https://openai.com/index/hello-gpt-4o/}}

\section{Conclusion and Future Work}
This technical report has introduced an efficient Vision-Language model, namely StrucTexTv3, which effectively tackles the problems of text-rich image perception and comprehension. The proposed combination of a multi-scale reduced visual transformer and MG-Sampler successfully solves the challenges of high-resolution input and rich representation learning for text-rich images. 
Subsequently, a systematic organization of tasks related to both perception and comprehension abilities of text-rich images, as well as high-quality data collection, was carried out to train a highly competitive unified large mulitmodal model. StrucTexTv3 stands out among numerous multimodal models equipped with LLM decoder of approximately 1.8B parameters, and it also makes the deployment of edge devices feasible.

Our method offers promising avenues for future research. (1) To enrich the comprehension of text-rich videos and multi-page scanned PDFs, we need to broaden the image-level context, enabling the extraction of visual tokens from multiple images. (2) To advance toward Artificial General Intelligence (AGI), we must augment our learning data with a wider range of images, corpora, and real user instructions. This will empower our model with more universal capabilities, such as GUI understanding and mathematical problem analysis. (3) To study the scaling laws of large vision-language models on text-rich images, we must validate the effectiveness of larger datasets and model parameter sizes. These efforts will pave the way for more comprehensive and intelligent text-rich image understanding in the future.


\bibliography{iclr2023_conference}
\bibliographystyle{iclr2023_conference}
\end{document}